\documentclass[10pt,twocolumn,letterpaper]{article}

\usepackage[pagenumbers]{cvpr} % To force page numbers, e.g. for an arXiv version

\usepackage[dvipsnames]{xcolor}

\definecolor{cvprblue}{rgb}{0.21,0.49,0.74}
\usepackage[pagebackref,breaklinks,colorlinks,citecolor=cvprblue]{hyperref}
\usepackage{multirow}
\usepackage{booktabs}
\usepackage{mathtools}
\usepackage{xcolor}
\usepackage{array}
\usepackage{marvosym}

\usepackage{multibib}
\newcites{appendix}{Appendix References}

\definecolor{lightblue}{rgb}{0.867, 0.917, 0.964}
\definecolor{ownorange}{rgb}{1.0,0.949, 0.8}

\newcommand{\bsmall}[1]{\textcolor{black}{\footnotesize{\texttt{#1}}}}

\newcommand{\authorskip}{\hspace{4.8mm}}
\def\paperID{5421} % *** Enter the Paper ID here
\def\confName{CVPR}
\def\confYear{2024}

\title{VL-GPT: A Generative Pre-trained Transformer\\ for Vision and Language Understanding and Generation}

\author{Jinguo Zhu\textsuperscript{1}\thanks{Equal contribution. This work is done when Jinguo Zhu is  an intern at Tencent AI Lab. The source code and model weights shall be released at \url{https://github.com/AILab-CVC/VL-GPT}. \textsuperscript{\Letter}Corresponding author.} \authorskip Xiaohan Ding\textsuperscript{2*} \authorskip Yixiao Ge\textsuperscript{2,3}\authorskip Yuying Ge\textsuperscript{2}\\
Sijie Zhao\textsuperscript{2}\authorskip Hengshuang Zhao\textsuperscript{4} \authorskip Xiaohua Wang\textsuperscript{1}\authorskip Ying Shan\textsuperscript{2,3\Letter} \\
{
\fontsize{10.4pt}{9.84pt}\selectfont
\textsuperscript{1} Xi'an Jiaotong University \hspace{5.5mm} \textsuperscript{2} Tencent AI Lab }\\ 
{
\fontsize{10.4pt}{9.84pt}\selectfont
\hspace{5.5mm} \textsuperscript{3} ARC Lab, Tencent PCG
\hspace{5.5mm} \textsuperscript{4} The University of Hong Kong}\\[2.5mm]
}

\begin{document}
\maketitle

\begin{abstract}

In this work, we introduce Vision-Language Generative Pre-trained Transformer (VL-GPT), a transformer model proficient at concurrently perceiving and generating visual and linguistic data.
VL-GPT achieves a unified pre-training approach for both image and text modalities by employing a straightforward auto-regressive objective, thereby enabling the model to process image and text as seamlessly as a language model processes text.
To accomplish this, we initially propose a novel image tokenizer-detokenizer framework for visual data, specifically designed to transform raw images into a sequence of continuous embeddings and reconstruct them accordingly.
In combination with the existing  text tokenizer and detokenizer, this framework allows for the encoding of interleaved image-text data into a multimodal sequence, which can subsequently be fed into the transformer model.
Consequently, VL-GPT can perform large-scale  pre-training on multimodal corpora utilizing a unified auto-regressive objective (\ie, next-token prediction).
Upon completion of pre-training, VL-GPT exhibits remarkable zero-shot and few-shot performance across a diverse range of vision and language understanding and generation tasks, including  image captioning, visual question answering, text-to-image generation, and more.
Additionally, the pre-trained model retrains in-context learning capabilities when provided with multimodal prompts.
We further conduct instruction tuning on our VL-GPT, highlighting its exceptional  potential for multimodal assistance.

\end{abstract}

\begin{figure*}
    \centering
    \includegraphics[width=0.95\linewidth]{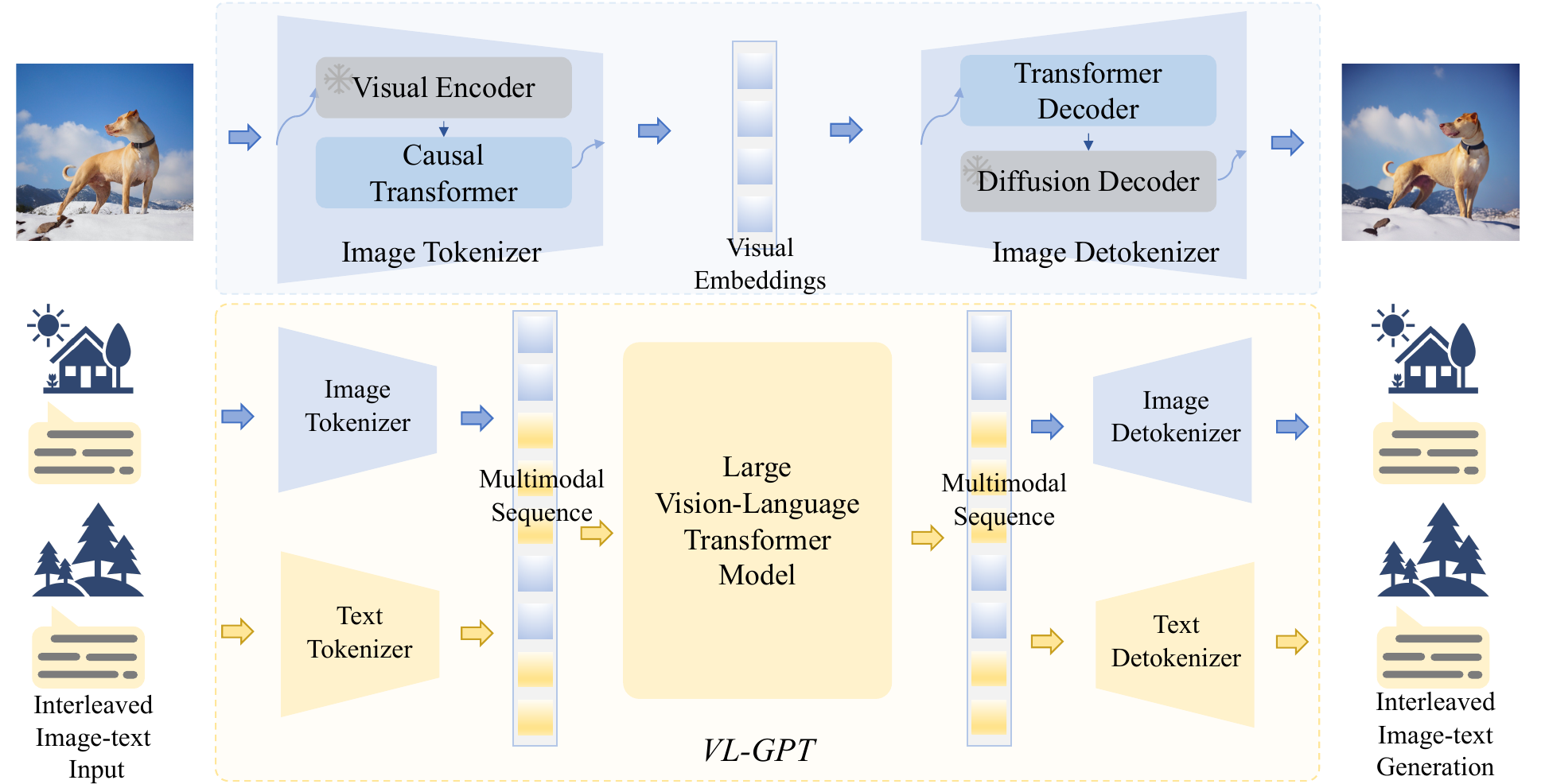}
    \caption{Overview of our proposed  approach.  The upper part delineates the image tokenizer-detokenizer framework, designed for encoding  images into continuous visual embeddings and reconstructing them in the pixel space. 
    The lower part  demonstrates the implementation  of our VL-GPT, where interleaved image-text data are encoded into multimodal sequence using  image and text tokenizers, subsequently processed by a  transformer model  auto-regressively. The image and text detokenizers are employed for generating respective outputs.}
    \label{overview}
    \vspace{-1.5em}
\end{figure*}

\section{Introduction}
\label{sec:intro}
Driven by the remarkable success of large language models (LLMs) in the field of natural language processing (NLP)~\cite{optllm,touvron2023llama,touvron2023llama2}, there has been a surge of interest within multimodal community to develop large vision-language (VL) models.
One of the promising approaches, exemplified by Flamingo~\cite{alayrac2022flamingo}, BLIP2~\cite{li2023blip2}, LLAVA~\cite{llava}, have explored how to build large VL models based on powerful pre-trained LLMs. 
These studies typically adopted a similar architecture: a  pre-trained image encoder and an LLM  are connected via a trainable connection module, which aligns the image feature and text embeddings, thereby enabling language models to accept  images and text as inputs and generate a text sequence.

To expand the capabilities of generating image in a multimodal context, certain efforts, \eg, Visual ChatGPT~\cite{wu2023visualchatgpt}, attempt to connect LLMs with image generation tools in a cascaded pipeline by transferring text messages, which inevitably introduce instability and noise.    
Alternatively, another line of research achieves it by optimizing models in an end-to-end manner~\cite{gill,dong2023dreamllm,wu2023nextgpt,pan2023kosmosg,li2023blipdiffusion}.
By aligning the output space with the image diffusion models, VL models can not only perceive but also  generate images and text.

A crucial characteristic of large language models is auto-regressive modeling~\cite{gpt1}, \ie, predicting next token, which facilitates language understanding and generation in a unified manner.
However, in the aforementioned studies, the inconsistency of image embeddings between LLM's input and output sides compels the model to treat input images and generated images differently, resulting in separate modeling for image understanding and generation.
Meanwhile, this discrepancy also  obstructs the implementation  of  auto-regressive training loss on image embeddings.

In this study, we introduce VL-GPT, a large vision-language generative pre-trained transformer that enables the unified training of both visual and linguistic data  using an auto-regressive objective, as depicted in Fig.~\ref{overview}.
To achieve this, we propose an  image tokenizer-detokenizer framework for the conversion between raw image pixels  and continuous visual embeddings, analogous to the role of the text tokenization~\cite{kudo2018sentencepiece,bpe} in language models.
The framework comprises an image tokenizer and an image detokenizer, where the tokenizer encodes raw images into a sequence of  continuous visual embeddings, and the detokenizer decodes the continuous  embeddings into pixel space.
To obtain visual continuous embeddings that are rich  in both image details and  semantic information,  we employ  the image embeddings and their corresponding caption embeddings extracted by pre-trained encoders (\ie, CLIP~\cite{clip}) as the supervision for  training of the framework.
Furthermore, the efficiency of the framework training is enhanced through weight initialization from pre-trained image encoders and high-quality image diffusion models.

By employing the image tokenizer-detokenizer framework,  visual embeddings can achieve consistency on both the input and output sides of the transformer model. 
Consequently, interleaved image-text data  can be trained in a unified auto-regressive manner. 
Specifically, the image tokenizer and the existing text tokenizer (\ie, BPE tokenizer~\cite{bpe}) first convert the image and text into a multimodal  sequence consisting of interleaved continuous visual embeddings and discrete text tokens.
The transformer  can then be trained to predict the next embedding or token in this multimodal sequence, employing  mean squared error (MSE) loss for continuous visual embeddings and cross-entropy loss   for discrete text tokens.
Contrary to previous works~\cite{gill,wu2023nextgpt,pan2023kosmosg,dong2023dreamllm}, all embeddings in the multimodal sequence can receive supervision from the auto-regressive loss.
During the generation stage,  visual embeddings and text tokens  can be generated auto-regressively without distinction, and subsequently decoded into raw images and text by the image detokenizer and text detokenizer, respectively.

Owing to the unified modeling, the pre-training of the VL model can be conducted on large-scale  image-text pairs and interleaved image-text data.
Upon completion of pre-training, the model is capable of perceiving arbitrary multimodal input  and generating  responses varying in modalities (\eg, text, images or their interleaved contents), allowing it to generalize to a wide range of vision and language understanding and generation tasks in a zero-shot or few-shot manner.
Moreover, the pre-trained model exhibits appealing emergent properties for multimodal in-context learning, as it can  effectively tackle  new unseen tasks when provided with multimodal prompts.
The  VL generative pre-trained transformer model, referred to as VL-GPT,  holds the potential to serve as a  powerful foundation model for the multimodal community, similar to the role of GPT family~\cite{gpt3,openai2023gpt4} in NLP. 
Our contributions are summarized as follows:
\begin{itemize}
    \item We propose an image tokenizer-detokenizer framework  to convert images into continuous embeddings and reconstruct them, while exploring effective training methods for this framework.  

    Through  efficient training that  requires an affordable computational cost, the image tokenizer and detokenizer can effectively retain both  semantic information and  pixel details of the original image.
    \item We introduce  VL-GPT, a generative pre-trained transformer model for vision and language (VL) understanding and generation tasks. The model can be pre-trained on large-scale multimodal corpora in a unified auto-regressive manner, \ie, predicting the next token in a multimodal sequence containing  continuous visual embeddings and discrete text tokens without any discrimination.
    \item VL-GPT exhibits  competitive performance on various VL understanding and generation   benchmarks under zero-shot and few-shot settings, including image captioning, visual question answering, and text-to-image generation. 
    It also demonstrates an appealing multimodal in-context learning ability when provided with multimodal prompts.
    Furthermore, it shows  promising potential to serve as a general multimodal assistant through instruction tuning.
\end{itemize}

\section{Related Work}
\label{sec:related}

\noindent \textbf{Multimodal Pre-training in the Pre-LLM Era.}
Prior research efforts primarily concentrated on model architecture to facilitate the fusion and interaction of cross-model data~\cite{chen2020uniter,yuan2021florence,yu2022coca}.
The success of transformers in language models~\cite{vaswani2017attention} and ViT~\cite{vit} inspired the development of unified multi-modal modeling~\cite{wang2022ofa,lu2022unifiedio}.
Although images and language can be processed by a unified model with shared parameters, they often have distinct training objectives.
It is worth mentioning that the BEiT series~\cite{bao2021beit,beit3} successfully adapted the masked language modeling objective from BERT~\cite{devlin2018bert} to  vision and multimodal pre-training.
\vspace{0.5em}

\noindent \textbf{Multimodal Pre-training in the LLM Era.}
Building upon pre-trained large language models (LLMs)~\cite{optllm,t5,touvron2023llama,touvron2023llama2},  recent studies have effectively developed multimodal language models capable of  processing image and text inputs to generate text outputs~\cite{li2023blip2,alayrac2022flamingo,llava,zhu2023minigpt,li2023otter}.
Another challenge for large multimodal models is generating multimodal content beyond language.
Several efforts, such as Visual ChatGPT~\cite{wu2023visualchatgpt} and HuggingGPT~\cite{shen2023hugginggpt}, have achieved this by connecting LLMs with other generation tools within an LLM integration framework, \eg, LangChain.
However, these systems exhibit instability and limited room for further optimization.
To enable LLMs to generate images with optimization, M-VADER~\cite{M-VADER} aligns the semantic consistence between an LLM and a diffusion decoder by training them on image-text pair data.
GILL~\cite{gill} achieves more complex interleaved image-text generation by mapping the embedding spaces of the LLM to text-to-image generation models.
NExT-GPT~\cite{wu2023nextgpt} extends this concept to additional modalities, such as audio and video.
DreamLLM~\cite{dong2023dreamllm} facilitates  passing the differential gradient from image diffusion models to language models, enabling the generation of free-form interleaved content.
Following similar methods, Kosmos-G~\cite{pan2023kosmosg} enhances the fidelity  of  generated images in context through a compositional instruction tuning task.

\begin{figure}
    \centering
    \includegraphics[width=0.95\linewidth]{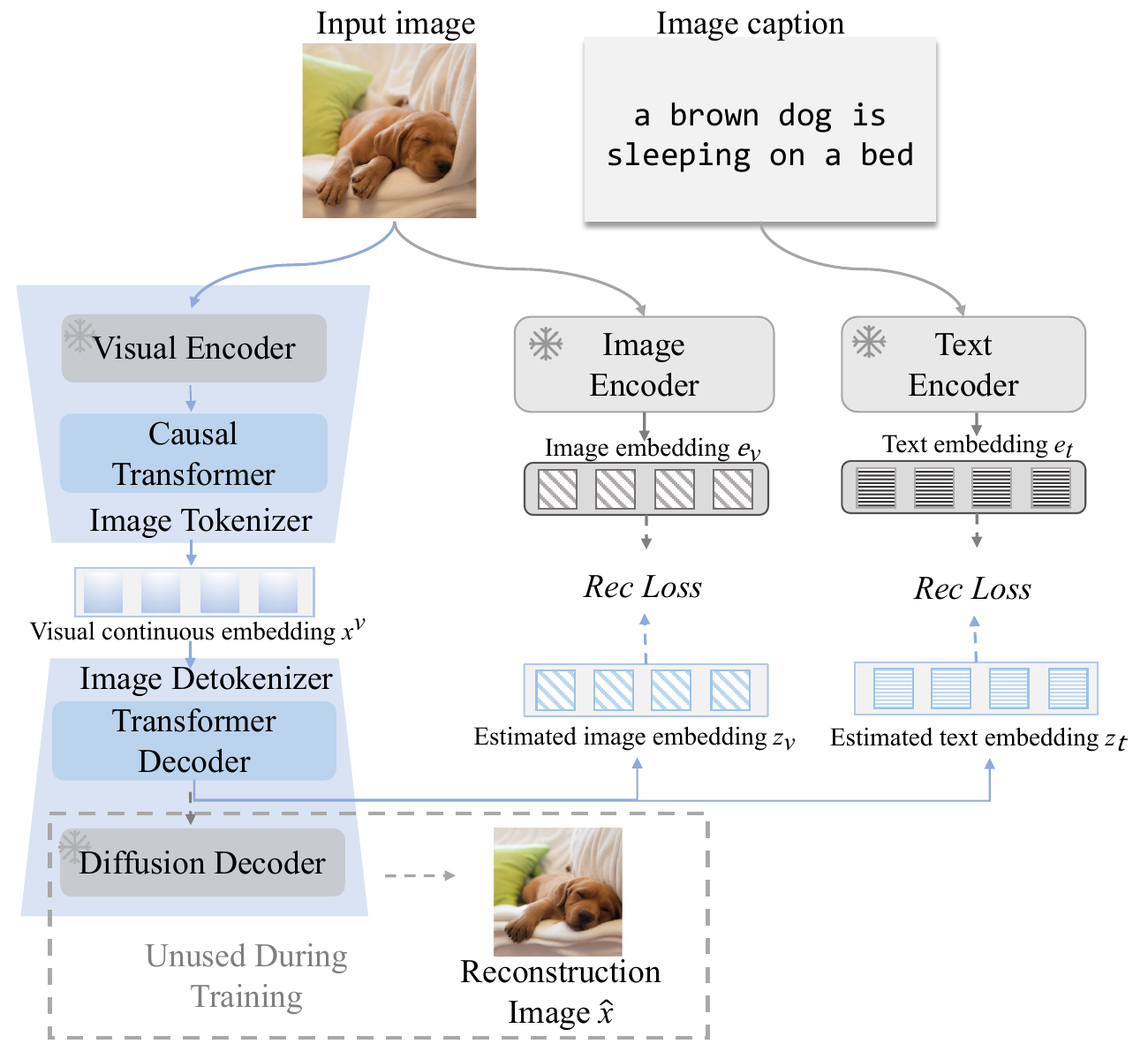}
    \caption{The training scheme of our image tokenizer-detokenizer framework, which is supervised by the frozen image and text encoders of our adopted pre-trained image diffusion model. Only the causal transformer in tokenizer and the transformer decoder in detokenizer necessitate training, while the diffusion decoder in detokenizer remains unused during training. }
    \vspace{-2em}
    \label{tokenizer_training}
    
\end{figure}

In contrast to our VL-GPT, these studies mainly focus  on leveraging existing LLMs and exploring the integration of current  image encoders and image generation models into LLMs. 
However, these methods do not achieve unified modeling for images and language, nor unified modeling for image understanding and generation.
For instance, special  queries are typically needed to encapsulate the context information for image generation, but they are deemed unnecessary when images serve as input for LLMs.
Moreover,  applying an auto-regressive training objective on visual embeddings is challenging due to the inconsistency of image embedding space.
Consequently, these approaches are limited in expanding the  scalable pre-training paradigm for the GPT family, \ie,  next-token prediction, to large vision-language models on web-scale multimodal corpora.

Recently, Emu~\cite{emu} proposes a multimodal pre-trained model that enables the auto-regressive training for both  visual and text embeddings.
However, it requires an costly second-stage fine-tuning of the Stable Diffusion~\cite{stablediffusion} to convert the visual embeddings into pixel space. In contrast,  our method utilizes a novel image tokenizer-detokenizer framework that can fully leverage a pre-trained image diffusion model (see Fig.~\ref{tokenizer_training}). This approach not only simplifies the process but also enhances training efficiency.
Similar to our approach, SEED~\cite{seed} initially trains an image tokenizer, followed by a multi-modal training.
Nevertheless, its tokenizer encodes images into discrete tokens via quantization operations, potentially losing partial image information.
In contrast, our tokenizer converts images into continuous  visual embeddings, preserving both semantic information and appearance details, resulting in improved performance across diverse benchmarks.

\section{Method}

As illustrated in  Fig. ~\ref{overview}, the implementation of our VL-GPT can be separated into two consecutive stages.
In the first stage, we learn an image tokenizer-detokenizer framework, capable of  encoding images into continuous visual embeddings and decoding them back.
The second stage is the pre-training and instruction tuning of our VL-GPT, which facilitates a unified modeling approach for vision and language understanding and generation.
In the following sections, we will provide a detailed description of these two stages.

\subsection{Image Tokenizer-Detokenizer Framework}
\label{tokenizerframework}
To implement an auto-regressive training objective on  visual embeddings and text tokens concurrently, we develop an image tokenizer-detokenizer framework for vision-language models.
The framework, inspired by text tokenizers utilized in language models~\cite{bpe}, 
can realize bi-directional conversion between original images and continuous visual embeddings, thereby enabling the transformer model to process vision data akin to processing text data.

\vspace{0.5em}

\noindent \textbf{Architecture}
The overall architecture of our image tokenizer-detokenizer framework is depicted in Fig. \ref{overview}.
It comprises two primary components: a tokenizer $\mathcal{E}$ responsible for encoding the image  into continuous visual embeddings, and a detokenizer $\mathcal{D}$ dedicated to decoding the visual embeddings back to raw images.

Formally, the image tokenizer $\mathcal{E}$  employs an image encoder (\eg, ViT~\cite{vit}) to extract spatial patched features $\boldsymbol{x}^p$ from the given image $\boldsymbol{x}$.
Subsequently, a standard decoder-only causal transformer is utilized to convert the patched features $\boldsymbol{x}^p$ to 1D (one-dimensional) visual embeddings  $\boldsymbol{x}^v \in \mathbb{R}^{N \times d}$, where  $N$ represents  the number of visual embeddings, and $d$ denotes the embedding dimension.
The 1D continuous visual embeddings $\boldsymbol{x}^v$ serve  as input embeddings to our vision-language model, analogous to word tokens in language models. 

Inspired by current image diffusion models with excellent performance and accessibility~\cite{stablediffusion,unclip,ip-adapter}, our image detokenizer $\mathcal{D}$ learns a latent diffusion model to decode visual embeddings $\boldsymbol{x}^v$ into images.
Specifically,  a transformer decoder is employed  to estimate condition  embedding $\boldsymbol{z}$  from $\boldsymbol{x}^v$.
Then a diffusion decoder, initialized from a pre-trained image diffusion models, can generate images $\boldsymbol{\hat{x}}$ based on  estimated condition embedding $\boldsymbol{z}$.

\vspace{0.5em}
\noindent \textbf{Training}
Despite the initialization with pre-trained models, conducting a full-scale end-to-end optimization of the image tokenizer and  detokenizer demands large-scale data and considerable training costs. 
To pursue efficient training,  
we opt to train the transformer decoder in image detokenizer to estimate the condition embedding utilized for the diffusion decoders, as illustrated  in Fig.~\ref{tokenizer_training}.
Notably, the diffusion decoder, including its U-Net and VAE modules, is not employed during framework training, substantially enhancing the efficiency of  training procedure.
 
As Fig. \ref{tokenizer_training} shows, the training objective of our framework aims to concurrently reconstruct the image condition embedding $e_v$ and text condition embedding $e_t$. This design distinguishes our framework from previous works~\cite{gill,wu2023nextgpt, seed}, which only align their intermediate  outputs with text embedding produced by the text encoder of the diffusion model. 
 Specifically, we optimize the framework by minimizing the following loss function (with weight $\lambda_1$ and $\lambda_2$):
\begin{equation}
\small
    L(\boldsymbol{z})= \lambda_1 * \operatorname{MSE}(z_v, e_v) + \lambda_2 * \operatorname{MSE}(z_t, e_t)
    \label{losstokenizer}
\end{equation}
 where $\operatorname{MSE}\left(\cdot\right)$ denotes the mean squared error loss, and $z_v$ and $z_t$ represent  the estimated image condition embedding and estimated text condition embedding, respectively.
 During inference, both types of condition embedding contribute collectively to  generate images.
Our image tokenizer-detokenizer framework can also work when reconstructing  only  image condition embedding (if $\lambda_2{=}0$) or only text condition embedding (if $\lambda_1{=}0$).
Moreover, the training for  estimating image embedding only  requires visual data, which is more training-friendly than estimating text embedding.
However, our experiments in Sec.~\ref{ablation} reveal that these two types of embedding complement each other: text embedding contain rich semantic information while image embedding effectively persevere image details.

\subsection{VL-GPT}

VL-GPT aims to process the vision and language understanding and generation within a single transformer model in a unified way, similar to GPT handles language tasks. 
It is capable of perceiving the interleaved multi-modal data and generating  content across various modalities.
By employing unified modeling, our VL-GPT can conduct auto-regressive pre-training on web-scale multimodal corpora, thereby holding the potential to serve as  a  powerful foundation model in the multimodal research community.

\vspace{0.5em}
\noindent \textbf{Architecture}
As depicted at the bottom of Fig. \ref{overview}, our VL-GPT comprises five components: a large vision-language transformer model $\mathcal{M}$,  an image tokenizer $\mathcal{E}_v$, 
a text tokenizer $\mathcal{E}_t$, an image detokenizer $\mathcal{D}_v$ and a text detokenizer $\mathcal{D}_t$.
In comparison to a language model, VL-GPT incorporates additional  image tokenizer and image detokenizer elements. 

Given any interleaved image-text data, the image tokenizer and the text tokenizer initially encode them into a multimodal sequence.
More specifically,  the image tokenizer $\mathcal{E}_v$   converts each image into $N$ continuous visual embeddings $\boldsymbol{x}^v$. 
Additionally, two special tokens $\mathtt{[IMG]}$ and  $\mathtt{[/IMG]}$ are appended at the beginning and end of the visual embeddings, respectively.
The visual embeddings are then combined with the discrete text tokens encoded by the text tokenizer  $\mathcal{E}_t$  to form a interleaved multimodal sequence  $\boldsymbol{v}=(v_1, v_2, \ldots, v_n)$, where $v_i$ can be either a discrete text token or a continuous  visual embedding. The multimodal sequence $\boldsymbol{v}$  is then fed into the large VL model $\mathcal{M}$ for unified auto-regressive modeling.

The output embedding $\mathcal{M}(v_i)$ can be flexibly transformed into a text embedding  through a language modeling head for the predefined vocabulary or into a visual embedding  with a separate regression head. 
During training, the selection of the transformed head depends on whether the  target for the current embedding is a  text token or a visual embedding.
During inference,  if $\mathtt{[IMG]}$ is predicted, the visual regression head will be utilized to transform  output embeddings in the subsequent $N$ prediction; otherwise, the language modeling head will be used.
The prediction embeddings are subsequently decoded to raw images or text via the image detokenizer $\mathcal{D}_v$  or the text detokenizer $\mathcal{D}_t$ .

\vspace{0.5em}

\noindent \textbf{Multimodal Pre-training.} Benefiting from the unified modeling of both visual and text embeddings, we can apply the unsupervised pre-training paradigm  of GPT~\cite{gpt1} to our VL-GPT on a large corpus of multimodal data with minimal modifications.

Given an interleaved multimodal sequence $\boldsymbol{v}=(v_1, v_2, \ldots, v_n)$ in a large-scale corpora, we employ the standard auto-regressive modeling objective in language models to maximize the following likelihood:
\begin{equation}
\small
L(\boldsymbol{v})=\sum_i^n \log P\left(v_i \mid v_1, v_2, \ldots, v_{i-1} ; \Theta\right)
\end{equation}
where $\Theta$ represents the parameters of our VL-GPT. 
We apply cross-entropy loss with a language modeling head on the discrete text tokens and utilize MSE loss with a regression head for continuous visual embeddings.

\vspace{0.5em}
\noindent \textbf{Instruction Tuning}
To enhance the ability of the pre-trained VL-GPT to  follow human instructions faithfully  and generate multimodal contents creatively, we perform further instruction tuning 
of VL-GPT using publicly available instruction-tuning datasets.
Briefly, the data from these datasets will be restructured into a conversational format, \ie, pairs of multimodal human instructions and their responses for single or multiple rounds, and subsequently employed for model tuning in a manner similar to the pre-training corpora.
A minor deviation  from pre-training process is that the training objective  will be applied exclusively to the embeddings tokenized from answer responses.

\section{Experiments}

The training of our VL-GPT consists of three phases: training for the tokenizer-detokenizer framework, unified multimodal pre-training for the vision-language transformer model, and instruction tuning for the pre-trained VL-GPT.
\begin{figure}[ht]
    \centering
    \includegraphics[width=0.98\linewidth]{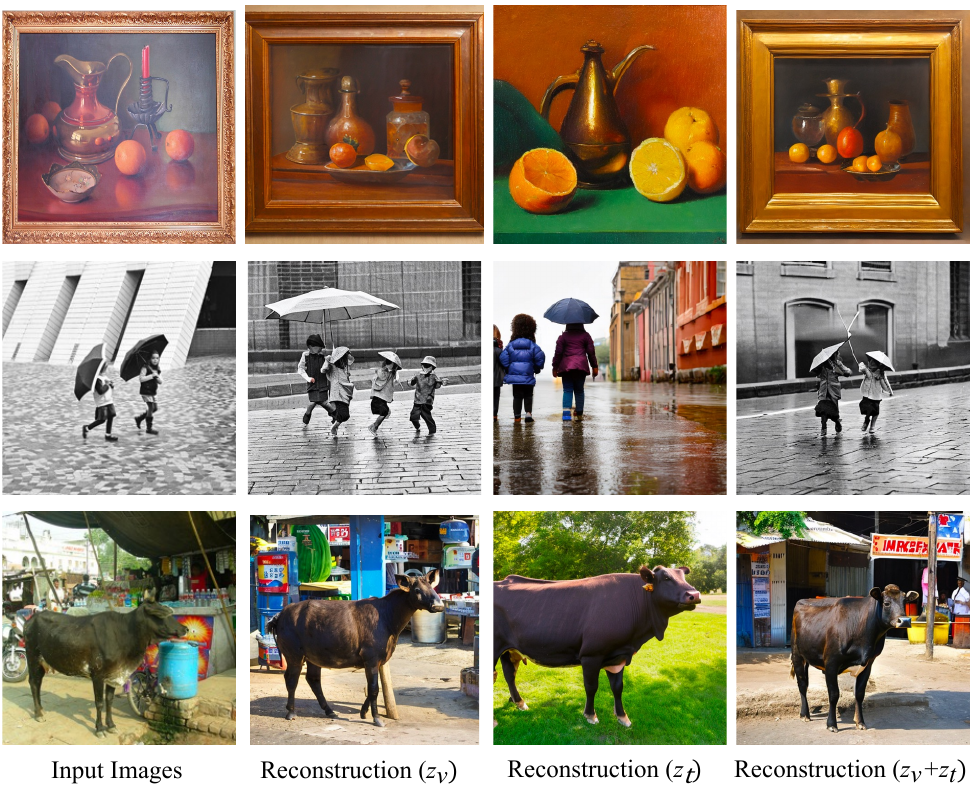}
    \vspace{-0.5em}
    \caption{Reconstruction images of our image tokenizer-detokenizer framework by utilizing  image condition embedding ($z_v$), or  text condition  embedding ($z_t$), or both types of condition embedding ($z_v{+}z_t$). More examples are included in the appendix.}
    \label{tokenizervis}
    \vspace{-1em}
\end{figure}

\begin{table}[t]
    \centering
    \small

\begin{tabular}{l|cc}
\toprule
Model & COCO & Flickr30k \\
\midrule
GILL~\cite{gill} & 67.45 & 65.16 \\
SD v1.5~\cite{stablediffusion} & 68.43 & 65.40 \\
SEED~\cite{seed} & 68.23 & 65.22 \\
unCLIP~\cite{unclip} & 79.30 & 
79.55 \\

\midrule
Our tokenizer-detokenizer & 80.22 &  79.14\\
\bottomrule

\hline
\end{tabular}

    \caption{ Evaluation of image reconstruction  with CLIP similarity.}
    \label{tokenizerclip}
    \vspace{-1em}
\end{table}

\begin{table*}[t]
\setlength{\tabcolsep}{6pt}
\renewcommand{\arraystretch}{0.9}
    \centering
   \small
   \resizebox{0.8\linewidth}{!}{
    \begin{tabular}{l|cccccc|c}
\toprule \multirow{2}{*}{ Models } & \multicolumn{6}{|c|}{ Image-Text understanding } & \multicolumn{1}{c}{ Text-to-image generations } \\
& COCO & VQAv2 & GQA & OKVQA & VizWiz & VisDial & COCO FID ($\downarrow$) \\
\midrule \midrule
\multicolumn{8}{l}{$\blacktriangleright$ \textit{VL Understanding or generation  Models}}  \\
\midrule

MetaLM~\cite{metallm} & 82.2 & 41.1 & - & 11.4 & - & - & -  \\
Kosmos-1~\cite{kosmos1} & 84.7 & 51.0 & - & - & 29.2 & - & - \\
Flamingo-9B$^\P$~\cite{alayrac2022flamingo} & 79.4 & 51.8 & - &44.7 & 28.8 & 48.0 & - \\
SD v1.5~\cite{stablediffusion} & - & - & - & - &- & - & 9.22 \\
\midrule
\midrule
\multicolumn{8}{l}{$\blacktriangleright$ \textit{Unified VL understanding and generation Pre-trained Models}}  \\
\midrule
GILL~\cite{gill} & - & - & - & - &- & - &  12.2 \\
Kosmos-G-1.9B~\cite{pan2023kosmosg} & - & - & - & - &- & - &  10.99 \\
SEED-OPT\textsubscript{2.7B}~\cite{seed} & 119.0 & 42.8 & 28.8 & - & - &- & -\\ 
 Emu~\cite{emu} & 112.4 & 52.0 & - & 38.2 & 34.2  &47.4& 11.66 \\
Emu$^\dag$~\cite{emu} & - & 52.9 & - & 42.8 & 34.4 & 47.8 & - \\
VL-GPT & 116.4 & 51.7 &34.6 &35.8 & 34.7 & 49.9  & 12.25\\
VL-GPT$^\dag$ & 119.2 & 55.3 & 38.1 & 41.5 & 35.2& 49.6 & -   \\
\midrule \midrule
\multicolumn{8}{l}{$\blacktriangleright$ \textit{Unified VL understanding and generation  Models with Instruction-tuning or Fine-tuning}}  \\
\midrule
CM3Leon-7B~\cite{CM3Leon} & 61.6 & 47.6 & - & 23.8& 37.6&22.6& 10.82 \\
Emu-I~\cite{emu} & - & 57.5 & - &46.2 & 38.1 & 50.1 & - \\
NExT-GPT$^{\S}$~\cite{wu2023nextgpt} & 156.7 & - & - & - & - & - & 11.28 \\ 
DreamLLM-7B~\cite{dong2023dreamllm}  & 115.4 & 56.6 & - & 44.3 & 38.1 &- & 8.46 \\
VL-GPT-I& 133.7 & 67.2 & 51.5 & 50.3 & 38.9 & 51.8  & 11.53 
\\

\bottomrule
\end{tabular}}

     \caption{Evaluation comparison between our VL-GPT and other models. $^\dag$ denotes that the zero-shot prompt is built by sampling two task-specific examples with their associated images removed. $^\S$ represents that the dataset employed for instruction tuning is private.}
    \label{zeroshotresults}
    \vspace{-2em}
\end{table*}

\subsection{Datasets}

 Publicly available datasets  are utilized for different  phrase of the VL-GPT training.
The image tokenizer-detokenizer framework is trained on image-text pairs from  CC3M~\cite{cc3m}, LAION-Aestheics~\cite{Laion-aesthetics}, and LAION-COCO~\cite{laioncoco}.
During the unified multimodal pre-training of VL-GPT,  a combination of  paired  and interleaved image-text  data is employed. The image-text pairs remain consistent with the  preview phase, while the interleaved 
image-text sequences are acquired from Multimodal-C4 (MMC4)~\cite{mmc4} and OBELICS~\cite{laurencon2023obelics}.  
We adopt similar preprocessing techniques for interleaved data  implemented in Flamingo~\cite{alayrac2022flamingo}. For each document, a maximum of  5 images and their associated captions are randomly sampled  to construct a subsequence with a token length of up to 512.
Additionally, for paired and interleaved image-text data, each image is randomly placed  before or after its corresponding caption.
For the instruction tuning of VL-GPT,  a compositional instruction tuning dataset is constructed from various sources, encompassing conversational data from LLAVA~\cite{llava} and SVIT~\cite{zhao2023svit},  image-text pair data from COCO Caption~\cite{cococap}, and image editing data from InstructPix2Pix~\cite{brooks2023instructpix2pix} and Magicbrush~\cite{zhang2023magicbrush}.
 These datasets are restructured into a conversational format using  the template provided in the appendix.
For further details regarding preprocessing and construction of our training dataset, please refer to the appendix as well.

\begin{figure*}[]
    \centering
    \includegraphics[width=0.98\linewidth]{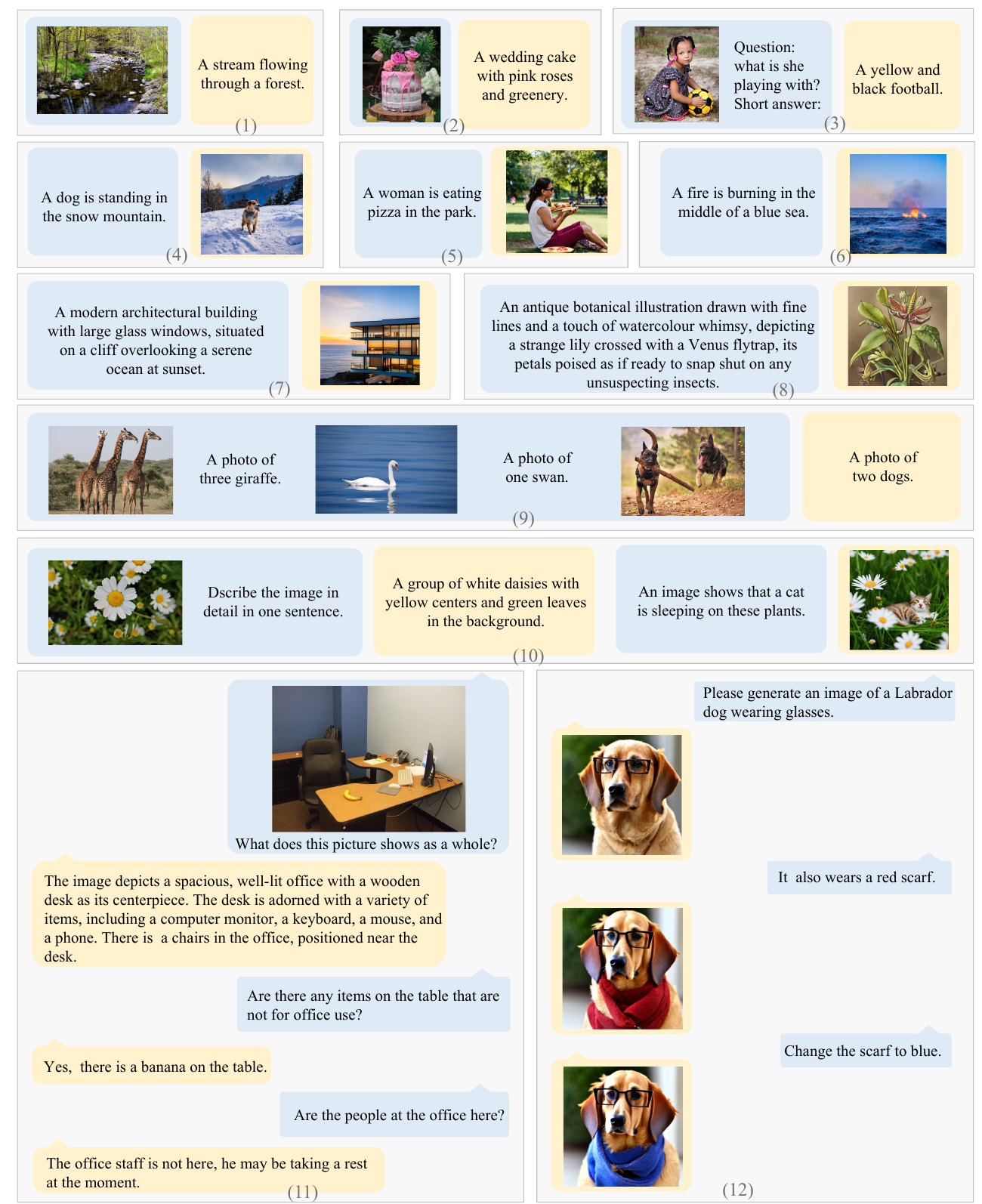}
    \caption{Illustrative samples of our VL-GPT across various  vision and language understanding and generation tasks.
    These tasks encompass: (1)-(2) image captioning, (3) visual question answering (VQA), (4)-(8) text-to-image generation, (9)-(10) multimodal in-context generation, and (11)-(12) multimodal dialogue.
    Examples (1)-(10) are generated by our pre-trained VL-GPT, while (11)-(12) are produced by our instruction-tuned VL-GPT.
    \colorbox{lightblue}{Blue boxes}  represent multimodal inputs and \colorbox{ownorange}{yellow boxes} indicate VL-GPT outputs.  }
    \label{visualization_total}
\end{figure*}

\subsection{Training setup}

To efficiently train the image tokenizer-detokenizer framework,  the visual encoder in the image tokenizer and the diffusion decoder in the image detokenizer are initialized with CLIP-L image encoder~\cite{clip} and IP-Adapter~\cite{ip-adapter}, respectively.
Moreover,  these two modules remain frozen throughout the entire training, and only the causal transformer and the transformer decoder necessitate optimization. 
Unless specified otherwise, the weight coefficients $\lambda_1$ and $\lambda_2$ in Eq.~\ref{losstokenizer} are assigned a value of 1.0 during both training and evaluation.
The AdamW optimizer~\cite{adamw} is employed for training, with a learning rate of 2e-4 and a cosine schedule.
 The framework is trained using a total batch size of 1024 on 8 NVIDIA 40G-A100 GPUs for 10,000 iterations. 

For the multimodal pre-training of our VL-GPT,  the pre-trained LLaMA 7B~\cite{touvron2023llama}, its text tokenizer,  and its text detokenizer are integrated with our trained image tokenizer and detokenizer to establish the VL-GPT model with a total of 7.5 billion parameters.
LoRA~\cite{hu2021lora} module is incorporated into the LLaMA model, resulting in  relatively low demand for computational resources.
 AdamW optimizer is also utilized with a learning rate of 2e-4.
The multimodal pre-training is conducted with a batch size of 4096 on 32 GPUs for 20,000 iterations.
% , taking about 3 days.
Instruction tuning is performed on the pre-trained VL-GPT, adopting  similar training settings used during pre-training.
 LoRA is also employed, and the learning rate is reduced to 5e-5.
The model is trained for 10,000 iterations with batch size of 512 on 4 GPUs.
% , requiring approximately one day.
Additional training settings are included in the appendix.

\subsection{Image Tokenizer and Detokenizer Performance}

The image tokenizer-detokenizer framework is designed to convert images  between pixel space and continuous visual embeddings.
To assess its  effectiveness,
we employ the method of calculating the CLIP similarity as the evaluation metric for our framework,  as implemented in SEED~\cite{seed}.
As demonstrated in Tab.~\ref{tokenizerclip},  our framework  achieves notably superior  semantic consistency compared to SEED, which utilized  quantized visual tokens.

Furthermore, we present visualizations of the reconstructed images generated by our framework in Fig.~\ref{tokenizervis}. 
By estimating both image condition embedding and text condition embedding and utilizing them to guide the  generation process of diffusion decoder, our image detokenizer is capable of generating images with high consistency in terms of spatial appearance and  semantic information.

\subsection{Evaluation of our VL-GPT}
\noindent \textbf{Benchmark Performance}
We  first evaluate the zero-shot performance  of VL-GPT on a variety of vision-language tasks, including image captioning on MSCOCO~\cite{cococap}, visual question answering on VQAv2~\cite{vqav2}, GQA~\cite{hudson2019gqa}, OKVQA~\cite{okvqa}, and VizWiz~\cite{vizwiz}, visual dialog on VisDial~\cite{visdial}, and text-to-image generation on MSCOCO.
Comprehensive details regarding these benchmarks and their metrics can be found in the appendix.
As  results in Tab.~\ref{zeroshotresults} indicate, VL-GPT  achieves competitive performance on both image-text understanding and text-to-image generation tasks, thereby validating  the effectiveness of unified multimodal pre-training.
Notably, VL-GPT attains an impressive CIDEr score of 116.4 or 119.2 on MSCOCO captioning without or with text-only prompts, surpassing other unified VL pre-trained models.
With further instruction tuning, VL-GPT-I, the instruction-tuned VL-GPT, significantly enhances model performance, achieving the best or near-best results in all tasks.

\begin{table}[t]
    \centering
    \small
    
\begin{tabular}{l|ccc|ccc}
\toprule
Models  & \multicolumn{3}{c}{ VQAv2 } & \multicolumn{3}{|c}{ VizWiz } \\
\quad $k$ & 2 & 4 & 8 & 2& 4 & 8  \\
\midrule

 Kosmos-1~\cite{kosmos1} & 51.4 & 51.8 & 51.4 & 31.4 & 35.3 & 39.0 \\
Flamingo-9B~\cite{alayrac2022flamingo} & - & 56.3 & 58.0 & - & 34.9 & 39.4 \\
Emu~\cite{emu} & 56.4 & 58.4& 59.0 & 37.8 & 41.3 & 43.9 \\

VL-GPT & 57.2 & 58.6& 58.9& 38.9 & 41.8& 44.2\\
\bottomrule
\end{tabular}

    \caption{Few-shot performance on visual  question answering.}
    \label{fewshot}
    \vspace{-1em}
\end{table}

\begin{table}[t]
\centering
\small
\begin{tabular}{@{}l|c|cc@{}}
\toprule
    Estimation target       & Reconstruction                               & \multicolumn{2}{c}{VL-GPT}                                                                               \\ \midrule
           & \begin{tabular}[c]{@{}c@{}}CLIP\\ Similarity ($\uparrow$)\end{tabular} & \begin{tabular}[c]{@{}c@{}}Captioning\\ CIDEr ($\uparrow$) \end{tabular} & \begin{tabular}[c]{@{}c@{}}Generation\\ FID ($\downarrow$)\end{tabular} \\
$e_t$       &   73.59 &    131.1        &    12.79    \\
$e_v$      &  80.05   &   123.6    &   13.61     \\
$e_t+e_v$ &  80.22   &   133.7         &   12.25   \\ 
\bottomrule
\end{tabular}
\caption{Ablation of condition embedding types. Text embedding ($e_t$), image embedding ($e_v$), or their combination ($e_t+e_v$) are employed to guide the training of the tokenizer-detokenizer framework. We evaluate the effectiveness of reconstructing images and the performance of VL-GPT when adopting different image tokenizer and detokenizer. }
\label{regresstarget}
\vspace{-1em}
\end{table}

\vspace{0.5em}
\noindent \textbf{Multimodal In-context Learning}
Similar to the behavior of LLMs, our  VL-GPT can be prompted to address new vision-language tasks when provided with a few  multimodal examples from training data composed in the multimodal prompt.
To quantitatively evaluate its multimodal in-context learning capability,
we examine the few-shot performance of VL-GPT when varying the number of examples in the given prompt, as shown in Tab.~\ref{fewshot}.
Our VL-GPT outperforms other works under almost all few-shot setting ($k{=}2,4,8$) on two datasets for the visual question answering task.
Moreover, a positive correlation is observed between the number of the examples in the given prompt and the performance on these two datasets.

\vspace{0.5em}

\noindent\textbf{Qualitative Results}
 Fig.~\ref{visualization_total} showcases a series of  generated visualizations using our VL-GPT model, encompassing various tasks such as image captioning, visual question answering, text-to-image generation, multimodal generation with in-context learning, and multimodal multi-turn dialogue.
Intriguingly, VL-GPT demonstrates remarkable capabilities that are not readily assessed through existing academic benchmarks.
For instance, in Fig.~\ref{visualization_total} (7-8), VL-GPT generates highly realistic images in response to long-text prompts containing complex concepts.
In Fig.~\ref{visualization_total} (10), VL-GPT exhibits the ability to generate images and texts in a  flexible manner, conditioned on the provided multimodal context. Fig.~\ref{visualization_total} (11-12) illustrates the multi-turn dialogue capabilities of the instruction-tuned VL-GPT, wherein the model generates multimodal contents consistent with the existing context based on user instructions.
This suggests the  promising potential of the VL-GPT as a versatile and effective  multimodal general assistant.

\subsection{Ablation Studies}
\label{ablation}
Previous studies typically generate images by converting their output into text condition embedding for image diffusion models.
In contrast,  our detokenizer estimates  both text condition embedding and image condition embedding from visual continuous embeddings, as depicted in Sec.~\ref{tokenizerframework}.
The advantage of this design will be discussed next.

Fig.~\ref{tokenizervis} displays the images reconstructed  by our tokenizer-detokenizer using different  estimated condition embedding, \ie, only using image condition embedding, only using text condition embedding, or using both. These examples reveal that these two type of embedding complement each other: image embedding effectively  preserve image appearance details while text embedding assists in image  reconstruction, \eg, determining the number of people.

As evidenced in Tab.~\ref{regresstarget},  although it is feasible to train image tokenizer-detokenizer framework by estimating solely one type of condition embedding (when $\lambda_1{=}0$ or $\lambda_2{=}0$ in Eq.~\ref{losstokenizer}), the simultaneous estimation of both types  of condition embedding leads to optimal performance  for both the tokenizer-detokenizer framework and VL-GPT.
We hypothesize that  estimating image condition embedding enables our tokenizer to retain more pixel information  from the input image, which is beneficial for image reconstruction. Meanwhile, estimating text condition embedding allows the visual embeddings to contain more high-level semantics, leading to improved performance in subsequent vision and language tasks.

\section{Conclusion}

We propose VL-GPT,  a generative pre-trained transformer model for vision and language understanding and generation.
The model incorporates an innovative image tokenizer-detokenizer framework, enabling it to be pre-trained on large-scale multimodal corpora with a unified auto-regressive objective.
Upon completion of the pre-training, VL-GPT exhibits competitive performance across various academic benchmarks and manifests several appealing emergent capabilities.
As for limitations, the effectiveness of our method has not been verified through the scaling up of model parameters.
We hope that our work will stimulate further exploration in the pursuit of  general intelligence within the multimodal research community.
{
    \small
    \bibliographystyle{ieeenat_fullname}
    \bibliography{main}
}

\clearpage

\section{Training Details}

\subsection{Training of image tokenizer and detokenizer}

\noindent \textbf{Datasets.}
The image-text pairs from CC3M, Laion-Aestheics, and LAION-COCO are utilized for the training of our image tokenizer-detokenizer framework.
Specifically,  CC3M dataset comprises  3.3 million  image-text pairs crawled from the Web.
Both Laion-Aesthetics and LAION-COCO  are  subsets of the larger LAION-5B dataset. 
LAION-Aesthetics is characterized by its high aesthetic quality, while  LAION-COCO is composed of images sampled from LAION-5B and their corresponding  captions generated  by existing vision-language models, \eg, BLIP.
Due to the efficient design of the framework, a relatively small subset of 10 million samples from these two datasets was found to be sufficient for model convergence in our experiments.
Further exploration of experiments with larger datasets remains a prospect for future research.
During the training process, data were randomly sampled  from the mixture  of these three datasets in a ratio proportional to their respective sizes.

\vspace{0.5em}
\noindent \textbf{Optimization.}
The visual encoder in the image tokenizer is initialized with CLIP-L, while the diffusion decoder in the image detokenizer incorporates the U-Net and VAE modules from IP-adapter Plus.
These components remain  frozen during the training process.
The  causal transformer in the image tokenizer and the transformer  decoder in the image decoder are constructed using    the standard transformer  decoder, which consisting of 12 transformer blocks with random initialization. 
Each block comprises a  causal self-attention layer, a cross attention layer, and a multilayer perception (MLP) layer.
The causal attention  layer plays a vital role in capturing causal dependencies among 1D visual continual embeddings, which is proved to be effective for further modeling in large vision-language models like Emu and SEED.
In all experiments, the number of visual embedding  $N$ is set to 32 in our all experiments,  and its dimension $d$ is set to 4096.  
 Image augmentation techniques employed in CLIP models are applied, which involve  resizing  the input image with its shorter side to 224 pixels and cropping the image to a fixed size of 224$\times$224 pixels.

\begin{table}[t]
\small
    \centering
     \resizebox{1.0\linewidth}{!}{
    \begin{tabular}{llll}
\hline Dataset & Task & Split & Metric \\
 COCOCap & Scene Description & test & CIDEr $(\uparrow)$ \\
 VQAv2 & Scene Understanding QA & test-dev & VQA acc. $(\uparrow)$ \\
 GQA  & Scene Understanding QA  &  test-dev & VQA acc. $(\uparrow)$ \\ 
OKVQA & External Knowledge QA & val & VQA acc. $(\uparrow)$ \\
 VizWiz & Scene Understanding QA & test-dev & VQA acc. $(\uparrow)$ \\
VisDial & Image Dialogue & val & NDCG $(\uparrow)$ \\
\hline COCO & Text-to-Image Generation & val test &FID $(\downarrow)$ \\
\hline
\end{tabular}}
    \caption{Summary of the evaluation benchmarks.}
    \label{benckmarks}
\end{table}

\subsection{Pre-training of VL-GPT}

\noindent \textbf{Datasets.}
In addition to the datasets utilized for  training the image tokenizer-detokenizer framework,  publicly available  interleaved image-text data, \ie, MMC4 and OBELICS, are employed  for the pre-training of our vision-language transformer.
 During pre-training, multimodal sequences with interleaved image-text data are obtained from these two datasets.
For MMC4,  the core split is used,  and  low-quality samples with a CLIP similarity between the image and its caption below 0.24 are filtered out.
 For OBELICS, a sequence comprising 512 tokens is randomly sampled based on the arrangement of image and text data within the original document.
  To augment the probability of procuring sequences containing multiple images, single-image sequences are discarded  with a likelihood of 0.8.
 Throughout the training process, these two datasets  maintain equivalent sampling  probabilities, as do  the sampling probabilities for datasets comprising image-text pairs  and datasets containing interleaved image and text data.

\vspace{0.5em}
\noindent \textbf{Optimization.}
The large vision-language model, VL-GPT, is constructed by integrating the pre-trained language model LLaMA 7B with our image tokenizer-detokenizer framework. 
LoRA modules are attached to all linear layers in the vision-language transformer, with a LoRA rank of 32.
An additional linear  head is employed  as a separate regression head to predict the subsequent visual continuous embedding for the current embedding.
During multimodal pre-training,  only the parameters in LoRA modules  and the regression head are tuned, while all  parameters of pre-trained LLaMA, image tokenizer, and image detokenizer remain frozen to reduce training costs.
The  data augmentation techniques used in the previous stage are also utilized in this phase.

\begin{table*}[ht]
\small 
    \centering
    \setlength{\tabcolsep}{6pt}
\renewcommand{\arraystretch}{1.3} %行距
   \resizebox{0.95\linewidth}{!}{
    \begin{tabular}{lm{2.5cm}m{11cm}}
\toprule
\multicolumn{2}{c}{ \normalsize{System Message}}    &
 \bsmall{\texttt{You are a helpful, respectful and honest assistant.  Always answer as helpfully as  possible, while being safe.  Your answers should not include any harmful, unethical, racist, sexist, toxic, dangerous, or illegal content. Please ensure that your responses are socially unbiased and positive in nature. } }\\ 
 \midrule

 \multirow{5}{*}{\normalsize{Conversation Template}}
 & Image captioning &   \bsmall{USER: Provide a brief description of the given image <image> ASSISTANT: <caption>.} \\

 & Image generation & \bsmall{USER: Create an image that visually represents the description: <caption>.   ASSISTANT: Here's the image: <image>} \\

 & Image editing &  \bsmall{USER:<image> <editing prompt>. ASSISTANT: Here is the edited image: <image> }  \\

\bottomrule
\end{tabular}
}
    \caption{ Summary of prompt templates employed in instruction tuning. The notation ``\bsmall{<image>}" will be replaced with the image data.  ``\bsmall{<caption>}'' and ``\bsmall{<editing prompt>}"  will be substituted with the corresponding caption and editing instruction, respectively.}
    \label{sft_template}
    \vspace{-0.5em}
\end{table*}

\begin{table*}[t]
\small 
    \centering
        \setlength{\tabcolsep}{6pt}
\renewcommand{\arraystretch}{1.3} %行距
    \resizebox{0.95\linewidth}{!}{
    \begin{tabular}{lm{2.2cm}m{12cm}}
\toprule 
\normalsize{Model} & \normalsize{Task}  &  \normalsize{Template} \\
\midrule
\multirow{5}{*}{\normalsize{VL-GPT}}& Image captioning &  \bsmall{<image> Please describe this image in detail in one sentence. It shows} \\
 & Image generation &  \bsmall{An image of \texttt{<caption>. [IMG]}}  \\
 & Image QA &  \bsmall{<image> Based on the image, <question>?  Short answer:} \\
 & Image dialog & \bsmall{<image> an image of <caption>.  Based on the image, <question\textsubscript{1}>? Short   answer: <answer\textsubscript{1}>. $\cdots$  Based on the image, <question\textsubscript{n}>? Short answer:}  \\ 
  \midrule
\multirow{6}{*}{\normalsize{VL-GPT-I}} & Image captioning & \bsmall{USER: Provide a brief description of the given image.<image> ASSISTANT:}  \\
   & Image generation & \bsmall{USER: Create an image that visually represents the description: <caption>.  ASSISTANT: }  \\
 & Image QA &  \bsmall{USER: answer the question with the shortest answer <question>? ASSISTANT:  } \\
 & Image dialog & \bsmall{USER: <image> ASSISTANT: an image of <caption>. USER: <question\textsubscript{1}>? ASSISTANT: <answer\textsubscript{1}>. $\cdots$ USER: <question\textsubscript{n}>? ASSISTANT: }\\

  \bottomrule
\end{tabular}
}
    \caption{Summary of the prompting template utilized during model evaluation.The terms ``\bsmall{<image>}" and ``\bsmall{<caption>}" shall be substituted with the corresponding image and its caption. Additionally, the notations ``\bsmall{<question\textsubscript{i}>}  and ``\bsmall{ <answer\textsubscript{i}>}" will be replaced with the \textit{i}-th question and answer pair in the dialogue. \bsmall{[IMG]} denotes the special token indicating the start of visual continuous  embeddings.}
    \label{prompttemplate}
    \vspace{-0.5em}
\end{table*}

\begin{figure*}
    \centering
    \includegraphics[width=1\linewidth]{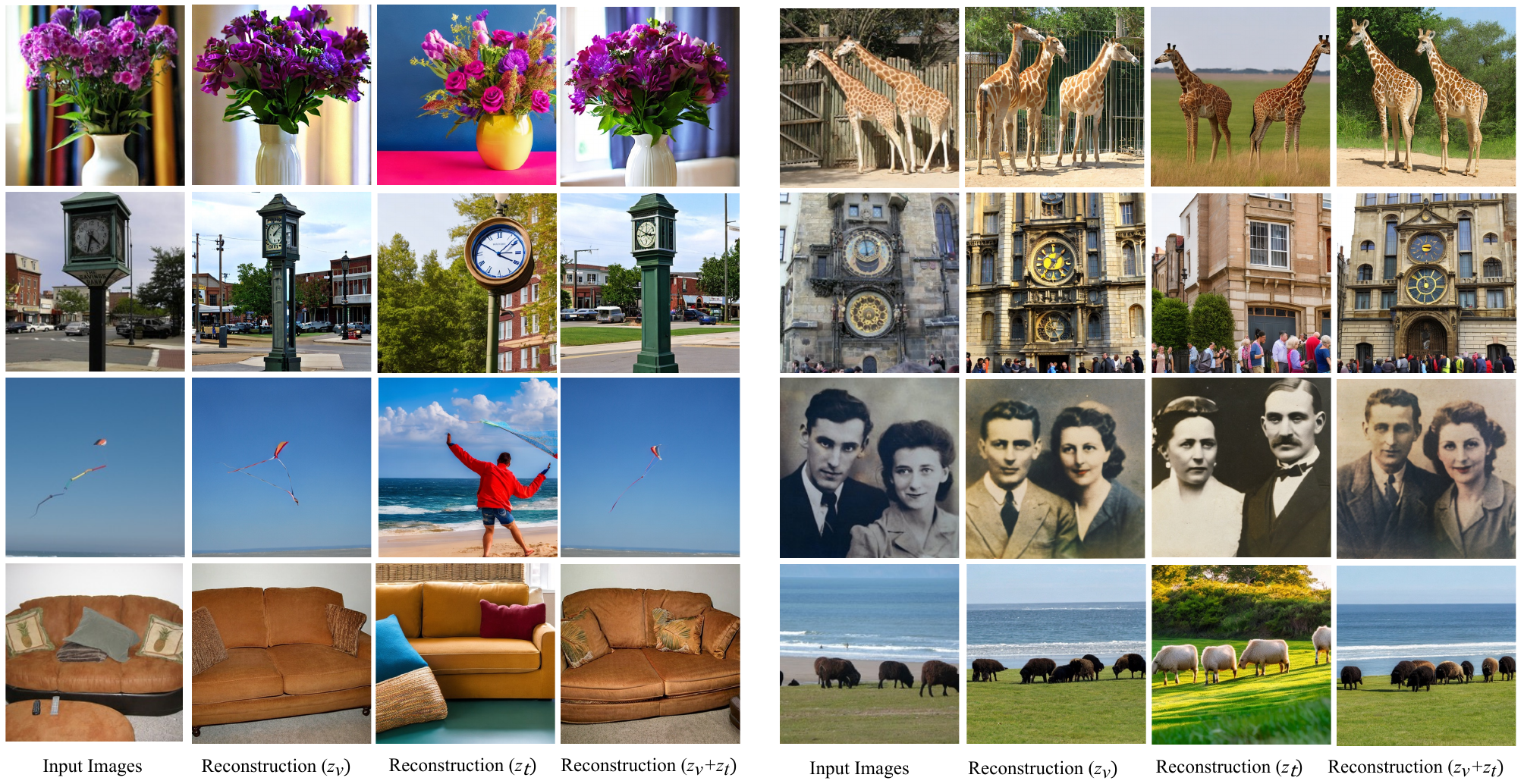}
    \caption{Reconstruction examples of our image tokenizer  and detokenizer by employing different condition embedding. }
    \label{tokenizervis2}
\end{figure*}

\begin{figure*}
    \centering
    \includegraphics[width=0.98\linewidth]{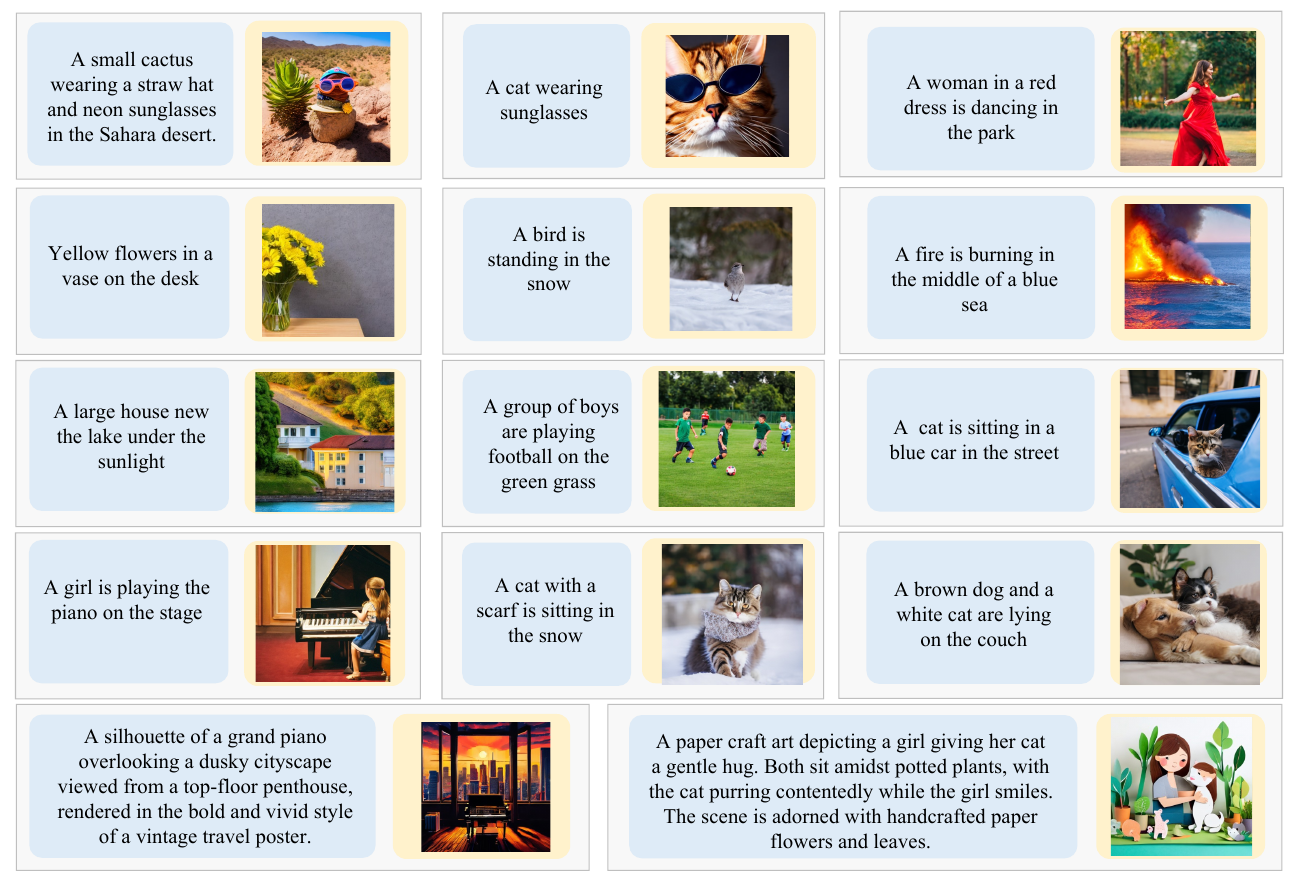}
    \caption{Examples of text-to-image generation.  \colorbox{lightblue}{Blue boxes}  denotes the text prompt, and \colorbox{ownorange}{yellow boxes} represents the generated image. }
    \label{text2imgvis2}
\end{figure*}

\begin{figure*}
    \centering
    \includegraphics[width=0.94\linewidth]{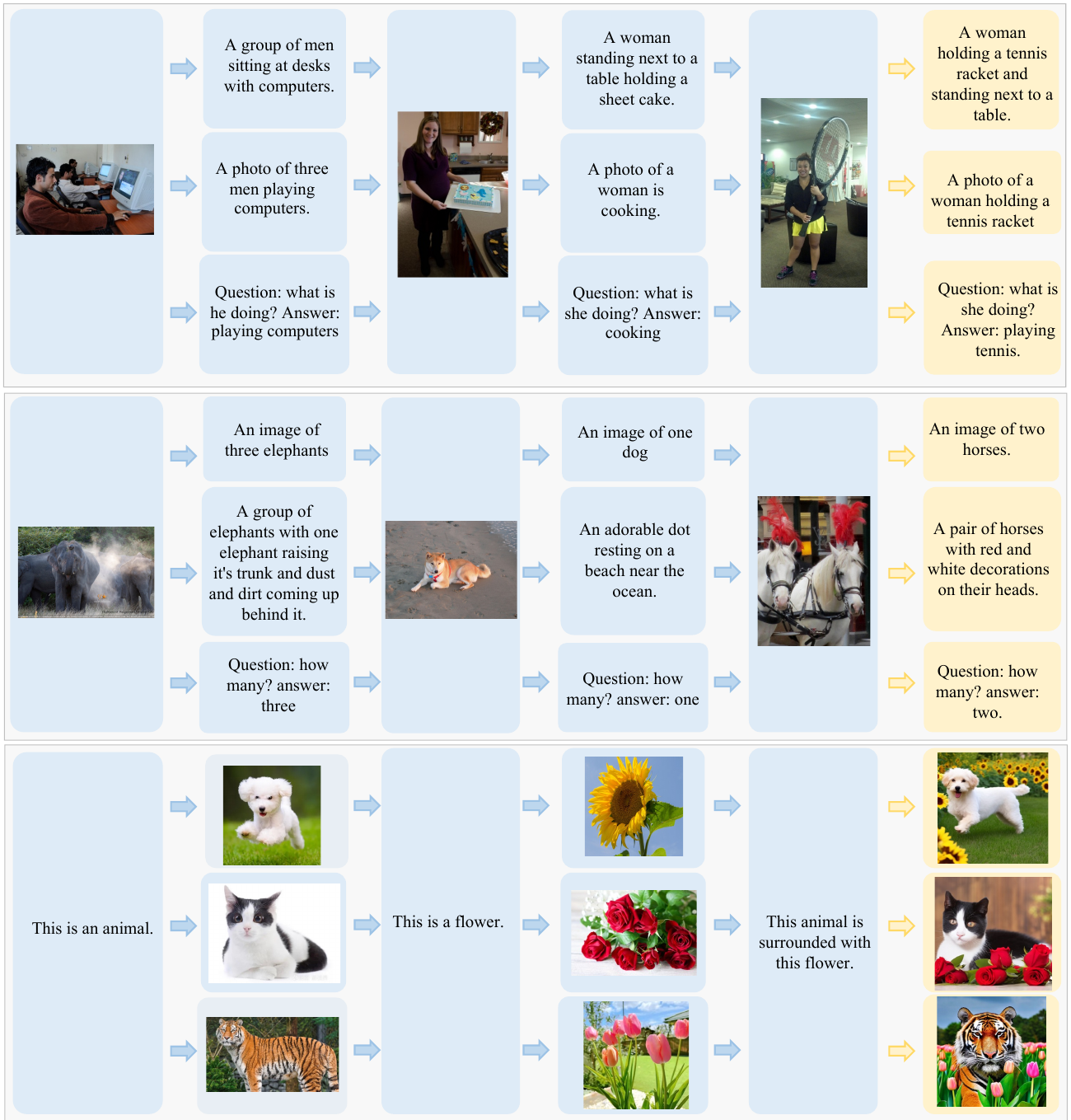}
    \caption{Three groups of  Multimodal In-context learning. 
    In each group,  partial content in the multimodal prompts  is changed, and our VL-GPT can generate desired outputs.  \colorbox{lightblue}{Blue boxes}  represent multimodal prompts and \colorbox{ownorange}{yellow boxes} indicate VL-GPT outputs. }
    \label{incontextlearningv2}
    \vspace{-0.5em}
\end{figure*}

\begin{figure*}
    \centering
    \includegraphics[width=1\linewidth]{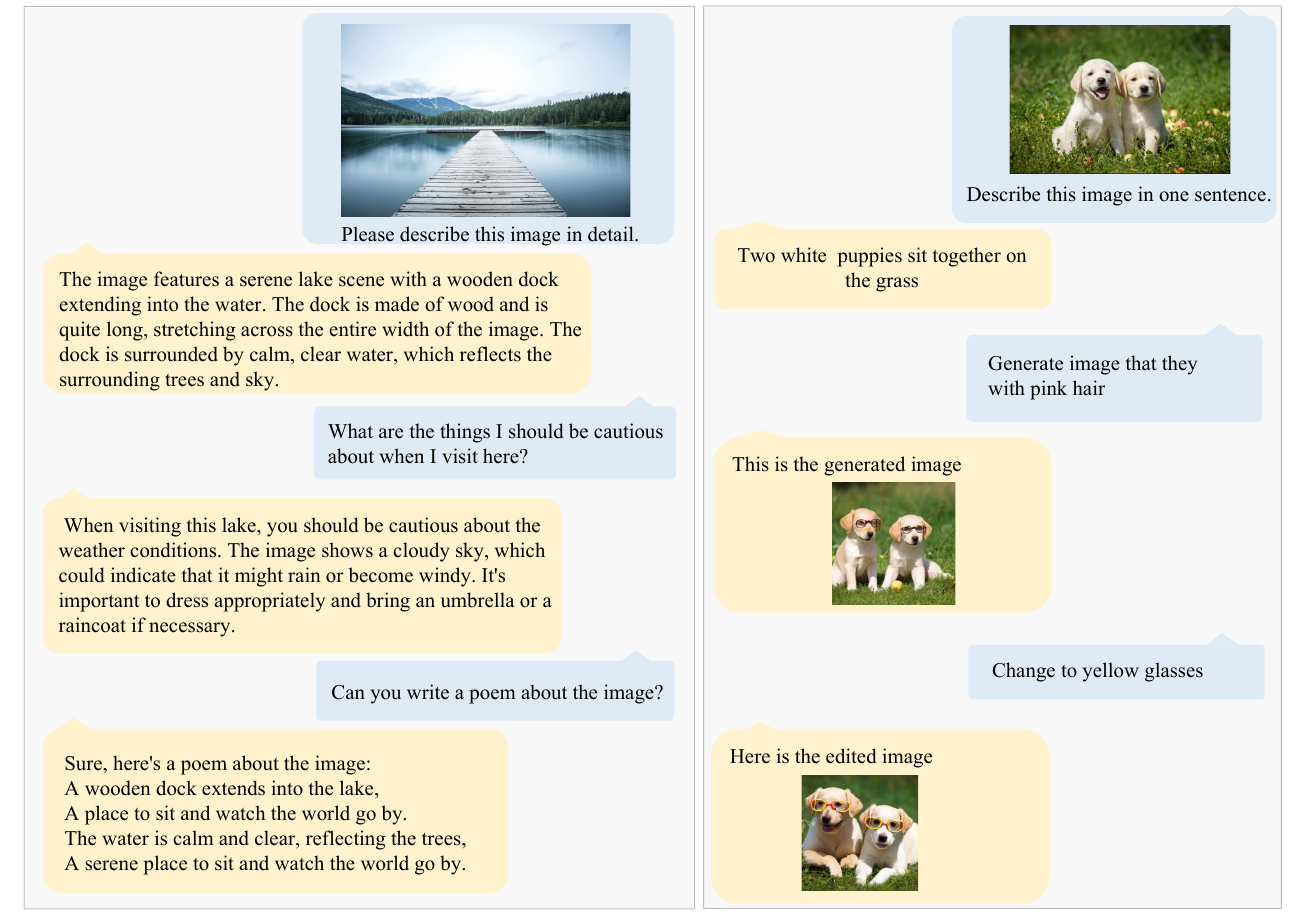}
    \caption{Qualitative examples of multimodal dialogue by our instruction tuned VL-GPT. \colorbox{lightblue}{Blue boxes}  denotes the user instructions, and \colorbox{ownorange}{yellow boxes} represents the assistant responses.}
    \label{dialogv2}
\end{figure*}

\subsection{Instruction tuning of VL-GPT}

\noindent \textbf{Datasets.}
To align the VL-GPT model with human instructions, multimodal instruction tuning is applied to the model using a combination of publicly available datasets, such as LLAVA, SVIT, MSCOCO Caption, InstructPix2Pix, and Magicbrush.
All dataset are restructured into a conversational formulation, consisting of a system message followed by a single-turn  or multi-turn conversation dialogue between a user and an assistant.
The system message and conversational template employed in our method are presented in Tab.~\ref{sft_template}.
Furthermore, the MSCOCO caption dataset is employed for both image captioning task and image generation task by altering the order of the image and its corresponding caption.
The InstructPix2Pix and Magicbrush datasets are utilized  for prompt-based image editing task.
During instruction tuning, data in these datasets are sampled  to construct a batch of data for model optimization in a ratio proportional to the dataset size. 

\vspace{0.5em}

\noindent \textbf{Optimization.}
Instruction tuning is carried out  on the pre-trained VL-GPT, with the training hyper-parameters  primarily  following those used in the pre-training phase.
As the training data for instruction tuning is significantly smaller than that employed for  pre-training, the batch size is set to a smaller number, \ie 512, and only four GPUs are utilized.
To prevent catastrophic forgetting of the pre-trained model,  the model is optimized with a reduced learning rate. 
Furthermore, LoRA modules are  applied in the transformer model, while all  other parameters remain frozen.

\section{Evaluation Details}
\subsection{Benchmarks}
To evaluate the vision and language understanding and generation 
 ability  of VL-GPT, we evaluate it on a variety of benchmarks, whose details and metrics are summarized in Tab.~\ref{benckmarks}.
Specifically, the test sample from any benchmark is first packaged with a task-specific prompt template and then tokenized into an incomplete multimodal sequence. Then the  VL-GPT model and its instruction tuned version, VL-GPT-I, are required to complete the multimodal sequence in an auto-regressive and open-ended manner. 
Evaluation results can be obtained by either using the official evaluation code or submitting our prediction on the official server.
It should be noted that not all results reported  in Tab.~\ref{zeroshotresults} are  zero-shot evaluation; for instance,  VL-GPT-I has been trained on COCO Caption.

\vspace{0.5em}
\subsection{Prompt Templates}
 To thoroughly capitalize on the  knowledge acquired during pre-training while  generating   outputs that adhere to the style of the benchmark under evaluation, we design task-specific prompt templates for the VL-GPT and VL-GPT-I.
 These templates are comprehensively outlined in  Tab.~\ref{prompttemplate}.

\vspace{0.5em}
\section{Qualitative Cases}
Additional reconstruction examples of our image tokenizer-detokenizer framework are illustrated in Fig.~\ref{tokenizervis2}.
Furthermore, the qualitative examples presented in
Fig.~\ref{text2imgvis2}, Fig.~\ref{incontextlearningv2}, and Fig.~\ref{dialogv2} demonstrate the superior performance of VL-GPT in various tasks, including text-to-image generation, multimodal in-context learning, and multimodal dialogue.

\end{document}